\documentclass[11pt]{article}

\usepackage{amsmath}
\usepackage{bm}
\usepackage{amsfonts}
\usepackage{enumerate}
\usepackage{tabularx}
\usepackage{makecell}
\usepackage{graphicx}
\usepackage{authblk}
\usepackage{natbib}
\usepackage{caption}
\usepackage{subcaption}
\usepackage{algorithm}
\usepackage{float}
\usepackage{listings}
\usepackage{dsfont}
\usepackage{booktabs}
\usepackage{algorithmic}
\usepackage{multirow}
\usepackage{booktabs}
\usepackage[colorlinks=true, linkcolor=blue, urlcolor=blue]{hyperref}

\title{Spatiotemporal Transformer for Imputing Sparse Data: A Deep Learning Approach}

\author[1]{Kehui Yao}

\affil[1]{Department of Statistics, University of Wisconsin-Madison}

\author[2]{Jingyi Huang}

\affil[2]{Department of Soil Science, University of Wisconsin-Madison}

\author[3]{Jun Zhu}

\affil[3]{Department of Entomology, Department of Biostatistics and Medical Informatics, Center for Demography and Ecology}

\date{}

\begin{document}

\maketitle

\begin{abstract}
	Effective management of environmental resources and agricultural sustainability heavily depends on accurate soil moisture data. However, datasets like the SMAP/Sentinel-1 soil moisture product often contain missing values across their spatiotemporal grid, which poses a significant challenge. This paper introduces a novel Spatiotemporal Transformer model (ST-Transformer) specifically designed to address the issue of missing values in sparse spatiotemporal datasets, particularly focusing on soil moisture data. The ST-Transformer employs multiple spatiotemporal attention layers to capture the complex spatiotemporal correlations in the data and can integrate additional spatiotemporal covariates during the imputation process, thereby enhancing its accuracy. The model is trained using a self-supervised approach, enabling it to autonomously predict missing values from observed data points. Our model's efficacy is demonstrated through its application to the SMAP 1km soil moisture data over a 36 x 36 km grid in Texas. It showcases superior accuracy compared to well-known imputation methods. Additionally, our simulation studies on other datasets highlight the model's broader applicability in various spatiotemporal imputation tasks.
\end{abstract}

\textbf{keywords}: Spatiotemporal Imputation, 
Transformer Model, Soil Moisture Data

\section{Introduction}
Soil moisture is essential for enhancing weather forecasts and understanding ecosystems. The NASA Soil Moisture Active Passive (SMAP) mission, launched in January 2015, was aimed to provide high-resolution soil moisture data beneficial for regional or local studies. However, a technical failure of the SMAP radar on July 7, 2015, led to a halt in the production of high-resolution soil moisture data. In the SMAP post-radar phase, various methods were explored to resume generating high-resolution soil moisture data. Research by \citet{das2019smap} suggested that combining Sentinel-1A and Sentinel-1B data with SMAP radiometer data in the active-passive algorithm has the potential to retrieve soil moisture at finer spatial resolutions of 1 and 3 km. However, this integration resulted in a decreased temporal resolution of SMAP active-passive data to a range of 3 to 12 days, which poses several challenges. These include limitations in monitoring short-term soil moisture changes, potential inaccuracies in assessing long-term soil moisture trends, and difficulties in real-time decision-making due to gaps in data.

To address temporal gaps in SMAP/Sentinel-1 data, filling missing values is a common approach.  \citet{kornelsen2014comparison} evaluated methods like monthly average replacement (MAR), soil layer relative difference (SLRD), linear interpolation, Evolutionary Polynomial Regression (EPR), and Artificial Neural Networks (ANN) for this purpose. They found these methods effective for minor data gaps but less so for larger gaps. \cite{park2023long} developed a Multilayer Perceptron (MLP) for long continuous gaps, but their approach is limited to single time series imputation at a specific location without considering spatial correlations. \citet{mao2019gap} used a two-layer machine learning-based framework for gap-filling the SMAP/Sentinel-1 3-km product. However, their method is oriented more towards the prediction of soil moisture using external covariates, which doesn't fully leverage the spatiotemporal structure within the data that requires imputation.

 Beyond the specific application for soil moisture, a wide range of time series imputation methods is available. Deep-learning-based imputation methods, in particular, have been increasingly popular for their effectiveness. These methods have shown enhanced performance compared to traditional techniques in several benchmark datasets, including those related to air quality \citep{yi2016st} and healthcare \citep{silva2012predicting}. The main strength of deep-learning approaches in time series analysis is their capability to identify and understand complex data patterns. 
 
 In the field of deep learning for time-series imputation, there are two main categories: autoregressive methods that use recurrent neural networks (RNNs), and non-autoregressive methods that utilize attention mechanisms. \citet{cao2018brits} introduced BRITS, an autoregressive method that uses bidirectional RNNs for imputing missing values, which was later improved by \citet{cini2021filling} through the integration of a graph neural network to enhance the understanding of spatial structures. RNN-based methods are known for their strong inductive bias, making them efficient even with smaller datasets. However, the training of RNNs can be slow, as they process data sequentially, and this might lead to the accumulation of errors over time, particularly in datasets with a significant number of missing values. On the other hand, attention-based methods, introduced by \citet{vaswani2017attention}, have demonstrated remarkable performance in numerous natural language processing tasks and offer an alternative to the recurrent structure for time series imputation. For instance, \citet{du2023saits} introduced SAITS, a model using self-attention mechanisms for filling in missing values in multivariate time series. \citet{tashiro2021csdi} proposed the idea of using a unique architecture of 2D attention (temporal attention and feature attention) for multivariate time series imputation. Compared to RNNs, Transformer-based models are better at capturing long-term dependencies and understanding global context, which is crucial for accurate imputation in time series with complex relationships. Also, transformer models avoid error propagation, are easier to optimize, and have a flexible attention scope, offering a more robust framework for handling sparse observations.

One limitation of the existing transformer-based imputation models is that they primarily focus on the data that requires imputation, often neglecting the potential benefits of incorporating additional contextual information. This means these models typically concentrate on the directly observed or partially missing spatiotemporal data, but they don't fully leverage other related data that could enhance the imputation process. Another limitation is that most models are not tailored for spatiotemporal data. They usually treat spatial information as features and use a full attention layer to learn the patterns, ignoring the common sense that nearby points are likely to be more similar, which is also known as the ``First Law of Geography" \citep{tobler1970computer}. Additionally, those models don't scale well when the spatial field is large.

To overcome the limitations of existing models, we develop a novel transformer-based model that has two key advantages. Firstly, it allows the integration of additional data sources for imputation. In the context of soil moisture, this model utilizes external covariates like daily surface weather data to help imputation. Secondly, our model uses a shifted-window-based self-attention mechanism for spatial modeling, which enhances the model's ability to capture local spatial correlations and scalability \citep{liu2021swin}.

The performance of our model was assessed through comparisons with different imputation methods, utilizing data from Texas soil moisture sites. In addition, we performed imputation tasks on various simulated datasets, such as the Healing Mnist \citep{krishnan2015deep} and the SMAP-Hydroblock \citep{vergopolan2021smap}. These simulation studies offer a further understanding of the model's effectiveness in a variety of scenarios.

This paper's contributions are twofold. Firstly, it introduces a self-supervised training framework built around a new spatiotemporal transformer model designed for a broad range of spatiotemporal imputation tasks. This model is notable for its ability to include covariates in the imputation process and its scalability to large spatial areas. Secondly, to our knowledge, this study is the first to use a transformer-based model for soil moisture gap-filling. In experiments conducted within the Texas region, the model outperformed various time-series imputation methods, achieving the lowest mean absolute error (MAE) and mean relative error (MRE).

\section{Method}\label{sec: method}
Let $S$ represent the discrete spatial domain and $T$ represent the discrete temporal domain, with $s \in S$ and $t \in T$. Define $L=|T|$ as the total number of time points and $K=|S|$ as the total number of spatial locations. The observation at spatial location $s$ and at time $t$ is denoted as $Y(s;t)$, and the entire spatiotemporal dataset is represented as $\boldsymbol{Y} \in \mathbb{R}^{K\times L}$.
 Let $\boldsymbol{M} \in \mathbb{R}^{K \times L}$ denote a binary mask, where each element corresponds to the presence or absence of a value in $\boldsymbol{Y}$. Specifically, if the observation $Y(s;t)$ is missing, then $M(s;t) = 0$; otherwise, $M(s;t) = 1$. Let $\boldsymbol{X} \in \mathbb{R}^{K \times L \times D}$ denote the external spatiotemporal covariates, where each element $X(s;t)$ comprises a vector of $D$ features. Furthermore, let $\tilde{\boldsymbol{Y}}$ denote the unknown complete data of $\boldsymbol{Y}$. The objective of spatiotemporal data imputation is to learn a model such that given inputs of $\boldsymbol{Y}$, $\boldsymbol{M}$, and $\boldsymbol{X}$, it yields an estimate $\hat{\boldsymbol{Y}}$ which minimizes the reconstruction error:
\begin{equation}
	L = \frac{\sum_{s=1}^K \sum_{t=1}^L l\{\hat{Y}(s;t), \tilde{Y}(s;t)\} \cdot \{1-M(s;t)\}}{\sum_{s=1}^{K}\sum_{t=1}^{L}\{1-M(s;t)\}},
\end{equation}
where $l(\cdot, \cdot)$ denotes an element-wise loss function such as mean absolute error (MAE). Given that $\tilde{\boldsymbol{Y}}$ is unknown, a surrogate objective is necessary. A typical approach is to create a set of artificially masked observations. This is done by randomly selecting a subset of the observed values and masking them, resulting in a conditional spatiotemporal dataset $\boldsymbol{Y}^{o}$ and the corresponding mask $\boldsymbol{M}^{o}$. The model then attempts to reconstruct $\boldsymbol{Y}$ based on $\boldsymbol{Y}^{o}$, and the surrogate loss is evaluated on the values that have been randomly masked, expressed by:
\begin{equation}
	L = \frac{\sum_{s=1}^K \sum_{t=1}^L l\{\hat{Y}(s;t), Y(s;t)\} \cdot \{M(s;t)-M^{o}(s;t)\}}{\sum_{s=1}^{K}\sum_{t=1}^{L}\{M(s;t)-M^{o}(s;t)\}}.
	\label{eq: loss function}
\end{equation}

\subsection{Model Overview}
\begin{figure}
\centering
\includegraphics[width=0.8\textwidth]{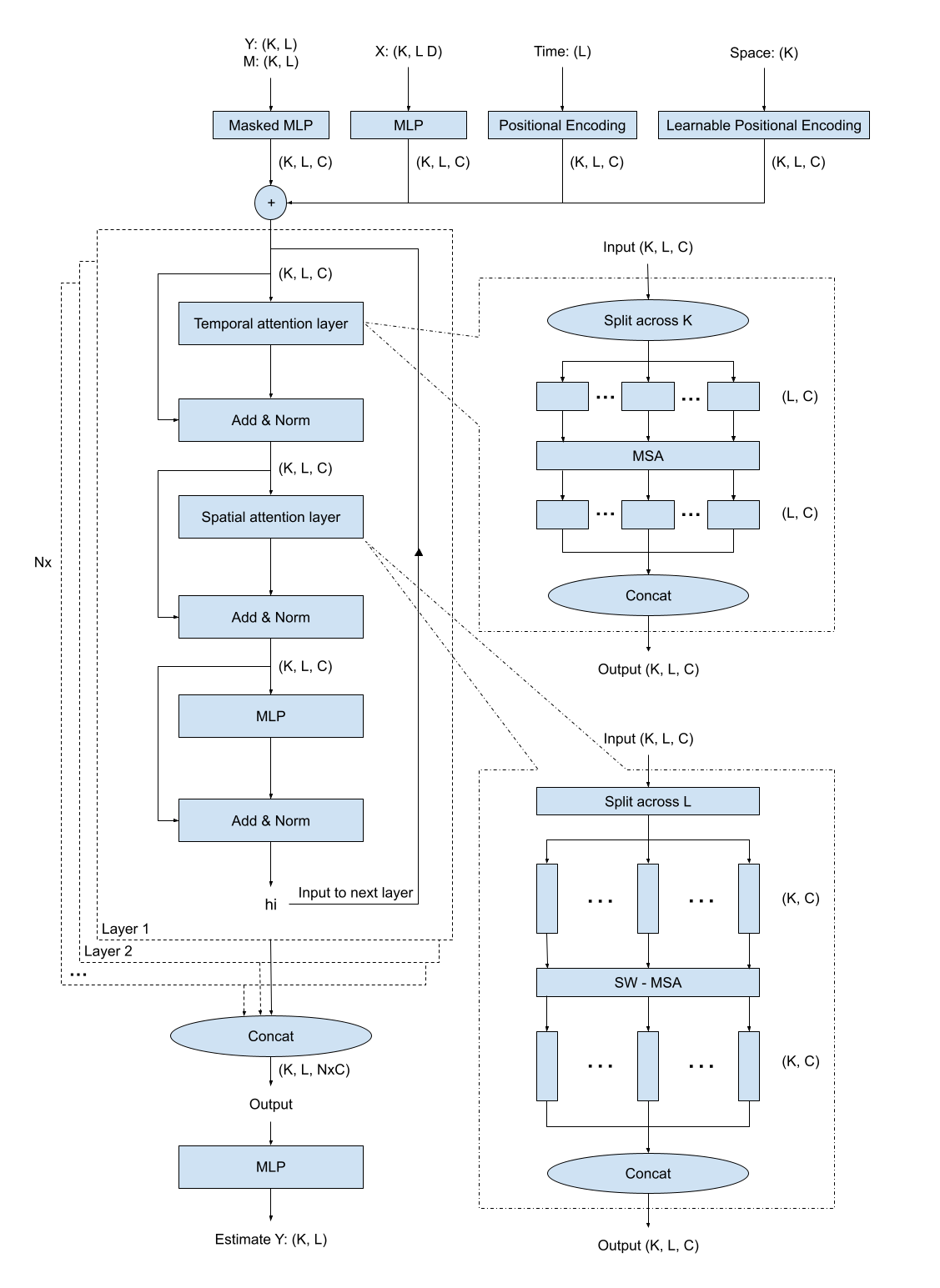}
\caption{The architecture of the Spatiotemporal Transformer model.}
\label{fig: st_transformer}
\end{figure}

In this section, we develop the Spatiotemporal Transformer model (ST-Transformer). The architecture of the model is depicted in Figure \ref{fig: st_transformer}. The ST-Transformer consists of three main components: an input encoder, a spatiotemporal transformer encoder, and an output layer. 

The input encoder functions by integrating covariate data ($\boldsymbol{X}$) and observed values ($\boldsymbol{Y}$) at each time point and location, transforming them into a latent representation. Following this, the spatiotemporal transformer encoder works to process the spatial and temporal dependencies present in the data. Finally, the output layer of the model is responsible for generating the imputed values.

\subsubsection*{Input Encoder} 
The input encoder consists of a Masked Multi-Layer Perceptron (MLP) to process the information from $\boldsymbol{Y}$, alongside a generic MLP for encoding the information in $\boldsymbol{X}$. It also consists of a positional encoding layer and a learnable spatial embedding layer to encode useful spatiotemporal positional information.

Mathematically, the process can be described as follows. Let $Y(s;t)$, $M(s;t)$, and $X(s;t)$ represent the observed data, missing data indicator, and predictive variables at location $s$ and time $t$ respectively. The masked MLP processes $Y(s;t)$ and $M(s;t)$ to produce an encoded representation $E_{Y}(s;t)$, formulated as: 

\begin{equation}
    E_{Y}(s;t) = \text{MLP}\{Y(s;t)\}\cdot M(s;t) + \text{mask-token} \cdot \{1-M(s;t)\},
\end{equation}
where $\text{mask-token}$ is a constant vector of dimension $C$. Similarly, the generic MLP processes $X(s;t)$ to yield an encoded representation $E_{X}(s;t)$, expressed as:
\begin{equation}
	E_{X}(s;t) = \text{MLP}(X(s;t)).
\end{equation}
The temporal position is encoded through a positional encoding layer following the approach in \citep{vaswani2017attention}. Specifically, the positional encoding for a position $t$ is given by:

\begin{align}
    E_{\text{time}2i}(t) &= \sin\left(t/10000^{\frac{2i}{C}}\right),\\
    E_{\text{time}2i+1}(t) &= \cos\left(t/{10000^{\frac{2i}{C}}}\right),
\end{align}
where the indices $2i$ and $2i+1$ represent the dimension indices of the positional encoding.
For spatial embedding, a learnable spatial embedding layer is used to obtain a representation $E_{\text{space}}(s)$ of the spatial location $s$:
\begin{equation}
	E_{\text{space}}(s) = \text{MLP}(s).
\end{equation}
The final unified latent representation $h(s;t)$ is then obtained by summing up these four encoded representations, as shown in the following equation:
\begin{equation}
	h(s;t) = E_{Y}(s;t) + E_{X}(s;t) + E_{\text{time}}(t) + E_{\text{space}}(s).
\end{equation}
 Therefore, $h(s;t)$ incorporates the essential information of the observed data and predictive variables, along with the spatiotemporal information.

\subsubsection*{Spatiotemporal Transformer Encoder}
The Spatiotemporal Transformer Encoder aims to encode valuable information for each data point by processing the spatial, temporal, and feature dimensions in the hidden data. It consists of multiple identical spatiotemporal attention layers, each connected via residual connections and layer normalization  \citep{ba2016layer}. Each layer comprises a temporal attention layer, a spatial attention layer, and an MLP, also linked through residual connections and layer normalization.


For the temporal attention layer, the input tensor is initially partitioned along the spatial axis, with each segment then processed through a multi-head self-attention layer (MSA) \citep{vaswani2017attention}. The processed segments are then merged to form the output of the temporal attention layer.

For the spatial attention layer, two variants are introduced. The first mirrors the temporal attention layer, differing only in the axis along which the input tensor is split and rejoined. However, for larger spatial dimensions, the conventional MSA encounters scalability issues. To mitigate this, the second variant adopts a stack of shifted-window-based multi-head self-attention (SW-MSA) \citep{liu2021swin} to replace the standard MSA to ensure efficient modeling.

MSA allows attention across varying positions within the input sequence by projecting the input data into multiple subspaces and performing self-attention in each of these subspaces simultaneously. Mathematically, the MSA layer is formulated as follows:

\begin{equation}
    \text{MSA}(x) = \text{Concat}(\text{head}_1, \text{head}_2, \ldots, \text{head}_h)W_O,
\end{equation}
where 
\begin{equation}
	\text{head}_i = \text{Attention}(xW_{Qi}, xW_{Ki}, xW_{Vi}),
\end{equation}
and
\begin{equation}
    \text{Attention}(Q, K, V) = \text{softmax}\left(QK^T/{\sqrt{d_k}}\right) V.
\end{equation}
Here, $W_{Qi}, W_{Ki}, W_{Vi}$ and $W_O$ are learned weight matrices, $h$ is the number of heads, $d_k=d_{model}/h$, where $d_{\text{model}}$ is the dimensionality of the input and output vectors, and $\text{Concat}$ denotes the concatenation operation. Each head computes self-attention independently over different projections of the input data, thereby allowing the model to learn and attend to different patterns across different subspaces. The outputs of all heads are then concatenated and linearly transformed to obtain the final output of the MSA layer. 

One limitation of MSA is its quadratic complexity with respect to the input length. For large spatiotemporal datasets, the spatial dimension often spans a magnitude of $10^2$ (representing a $10\times 10$ grid) to $10^4$ (representing a $100 \times 100$ grid), making MSA unscalable. Additionally, MSA computes global attention over the entire input sequence, neglecting the localized spatial correlations inherent in geospatial data. 

\begin{figure}
\centering
\includegraphics[width=\textwidth]{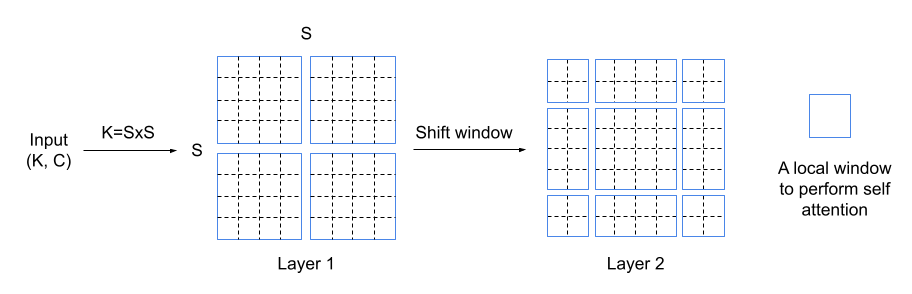}
\caption{Illustration of the shifted-window multi-head self-attention (SW-MSA) mechanism.}
\label{fig: sw_msa}
\end{figure}

SW-MSA addresses the computational challenge by dividing the large spatial field into non-overlapping smaller windows, within which self-attention is computed independently within each window \citep{liu2021swin}. To enable information exchange between windows, a shifted window strategy is utilized after initial window attention. This strategy re-partitions the data based on a shifted window, creating connections between previously non-overlapping windows, and computes masked attention within each window. It's common to stack multiple SW-MSA layers with different window shift sizes to capture spatial correlations at varying scales. The high-level picture of SW-MSA is illustrated in Figure \ref{fig: sw_msa}. In summary, shifted-window-based MSA balances local and global attention and is more scalable compared to traditional MSA. In certain configurations, the SW-MSA can be adjusted to function as the standard MSA. Unless specified otherwise, we default to using SW-MSA in the spatial attention layer in subsequent text.

Now, we demonstrate how the input tensor $\boldsymbol{h}$ traverses through a spatiotemporal attention layer as follows:

\begin{align}
		&\boldsymbol{h}_{\text{time}} = \big[\text{MSA}\{\boldsymbol{h}(1,\cdot)\}, \text{MSA}\{\boldsymbol{h}(2,\cdot)\}, \ldots, \text{MSA}\{\boldsymbol{h}(K,\cdot)\}\big]^{\prime}\\
		&\boldsymbol{h}= \text{LayerNorm}(\boldsymbol{h}+ \boldsymbol{h}_{\text{time}})\\
		&\boldsymbol{h}_{\text{space}} = \big[\text{SW-MSA}^n\{\boldsymbol{h}(\cdot, 1)\}, \text{SW-MSA}^n\{\boldsymbol{h}(\cdot, 2)\}, \ldots, \text{SW-MSA}^n\{\boldsymbol{h}(\cdot, L)\}\big]^{\prime}\\
		&\boldsymbol{h}= \text{LayerNorm}(\boldsymbol{h}+ \boldsymbol{h}_{\text{space}}).\\
		& \boldsymbol{h} = \text{LayerNorm}\{\boldsymbol{h} + \text{MLP}(\boldsymbol{h})\}
\end{align}

Here, $\text{SW-MSA}^n(x)$ is a stack of SW-MSA followed by residual connections and layer normalization. The mathematical representation is:

\begin{align}
\text{SW-MSA}^n(x) = \text{LayerNorm}(&x + \text{SW-MSA}_{n}( \\
&\text{LayerNorm}(x + \text{SW-MSA}_{n-1}( \\
&\ldots \text{LayerNorm}(x + \text{SW-MSA}_{1}(x))\ldots )))),
\end{align}
where $\text{SW-MSA}_i(x)$ is the $i$-th SW-MSA layer. 

Suppose our encoder consists of $N$ identical spatiotemporal attention layers. Dnoting the operation of a single layer (as outlined in the preceding procedure) as $L(\cdot)$. If we represent the input to the $i$-th layer as $\boldsymbol{h}_i$, then the input to the $(i+1)$-th layer can be expressed as:

\begin{equation}
    \boldsymbol{h}_{i+1} = \boldsymbol{h}_i + L(\boldsymbol{h}_i).
\end{equation}

Even though a single layer might suffice for simpler tasks, employing multiple layers could potentially improve performance. This enhancement arises as each additional layer provides the model an opportunity to learn more abstract representations of the input data. The final output from the encoder is formed by concatenating the outputs from all spatiotemporal attention layers.

\subsubsection*{Output Layer}

The Output Layer of our model is a generic MLP, which aggregates the output from the spatiotemporal transformer encoder to produce the final imputation. Firstly, hidden representations from spatiotemporal encoder $\boldsymbol{h}_i$, are concatenated along the feature dimension, formulated as:

\begin{equation}
\boldsymbol{H} = \text{Concat}(\boldsymbol{h}_1, \boldsymbol{h}_2, \ldots, \boldsymbol{h}_N).\end{equation}
Subsequently, the concatenated representation $\boldsymbol{H}$ is processed through a generic MLP to generate the final imputation, expressed as:
\begin{equation}
\hat{\boldsymbol{Y}} = \text{MLP}(\boldsymbol{H}).
\end{equation}

\section{Case Study: Soil Moisture Imputation}
\subsection{Remote-Sensored and Ground-Based Datasets}

\subsubsection*{Region of Interest}

The Texas Soil Observation Network (TxSON) is a well-monitored area covering 1300 $\text{km}^2$ near Fredericksburg, Texas, in the central part of the Colorado River \citep{caldwell2019texas}. It has 40 soil moisture monitoring stations placed at distances of 3, 9, and 36 km within the Equal-Area Scalable Earth Grid. We selected a region of 36x36 km by setting the boundaries based on the furthest longitudinal and latitudinal coordinates of all the in situ sites within TxSON. This region was then split into a 36x36 grid, where each cell covers an area of about 1 km x 1 km. 
In the following sections, we will describe the datasets used for our study, all of which were downloaded for this specific region for the time period from 2016-01-01 to 2022-12-31.

\subsubsection*{SPL2SMAP\_S}
Our initial dataset is SMAP/Sentinel-1 L2 Radiometer/Radar 30-Second Scene 3 km EASE-Grid Soil Moisture, Version 3 (SPL2SMAP\_S) \citep{das2019smap}. This Level-2 (L2) product provides soil moisture estimates at a 1km spatial resolution, based on the cylindrical EASE-Grid 2.0 projection \citep{brodzik2012ease}. We will refer to this dataset as SMAP 1km in subsequent text. However, the temporal coverage of this dataset is limited due to the less frequent revisit of the Sentinel-1 satellite, which has a revisit frequency of approximately 12 days. This infrequent revisit rate results in temporal gaps in the dataset, which is exactly what we aim to impute. Due to the low temporal resolution of this product, the resulting data contains around 95\% missing values.

\subsubsection*{SPL3SMP}
Our second dataset is SMAP L3 Radiometer Global Daily 36 km EASE-Grid Soil Moisture, Version 8 (SPL3SMP), accessible at \url{https://nsidc.org/data/spl3smp/versions/8} \citep{o2021smap}. Within our study region, this Level-3 (L3) product offers a close to daily soil moisture estimate at a 36km spatial resolution under the cylindrical EASE-Grid 2.0 projection \citep{brodzik2012ease}. We will refer to this dataset as SMAP 36km in subsequent text.

\subsubsection*{Daily Surface Weather Data}
Our third dataset is Daymet Version 4 \citep{thornton1840daymet}. This dataset offers daily weather parameters such as precipitation, solar radiation, minimum and maximum daily temperature, and vapor pressure at 1km spatial resolution. 

\subsubsection*{Soil, Topography, and Land Cover Type Data}
The other datasets utilized in our study are listed as follows:
\begin{description}
\item \textbf{Soil Texture Data:} We obtain the soil texture data from \citet{poggio2021soilgrids}, which provides data at a spatial resolution of 250 meters \citep{hengl2018clay}. These datasets provide information on the fractions of clay and sand, soil organic carbon content, and bulk density. We aggregate this data to a coarser spatial resolution of 1km.

\item \textbf{Elevation Data:}  We obtain elevation data from the U.S. Geological Survey's (USGS) Shuttle Radar Topography Mission (SRTM) Global 1 arc-second dataset, which has a spatial resolution of about 30 meters \citep{farr2007shuttle}. We also aggregate this data to a 1km spatial resolution.
    
    \item \textbf{Land Cover Type Data:} We obtain land cover type data from NASA's Moderate Resolution Imaging Spectroradiometer (MODIS) instrument, specifically the MCD12Q1 dataset \citep{friedl2022modis}. This dataset provides classifications of land cover into various categories including forest, urban, and water, among others. We specifically select the International Geosphere-Biosphere Programme (IGBP) classification band, which represents a particular classification scheme of land cover, to match the 1km spatial resolution of our study grid.
\end{description}


\subsubsection*{Visualization}
\begin{figure}[H]
\centering
\includegraphics[width=\textwidth]{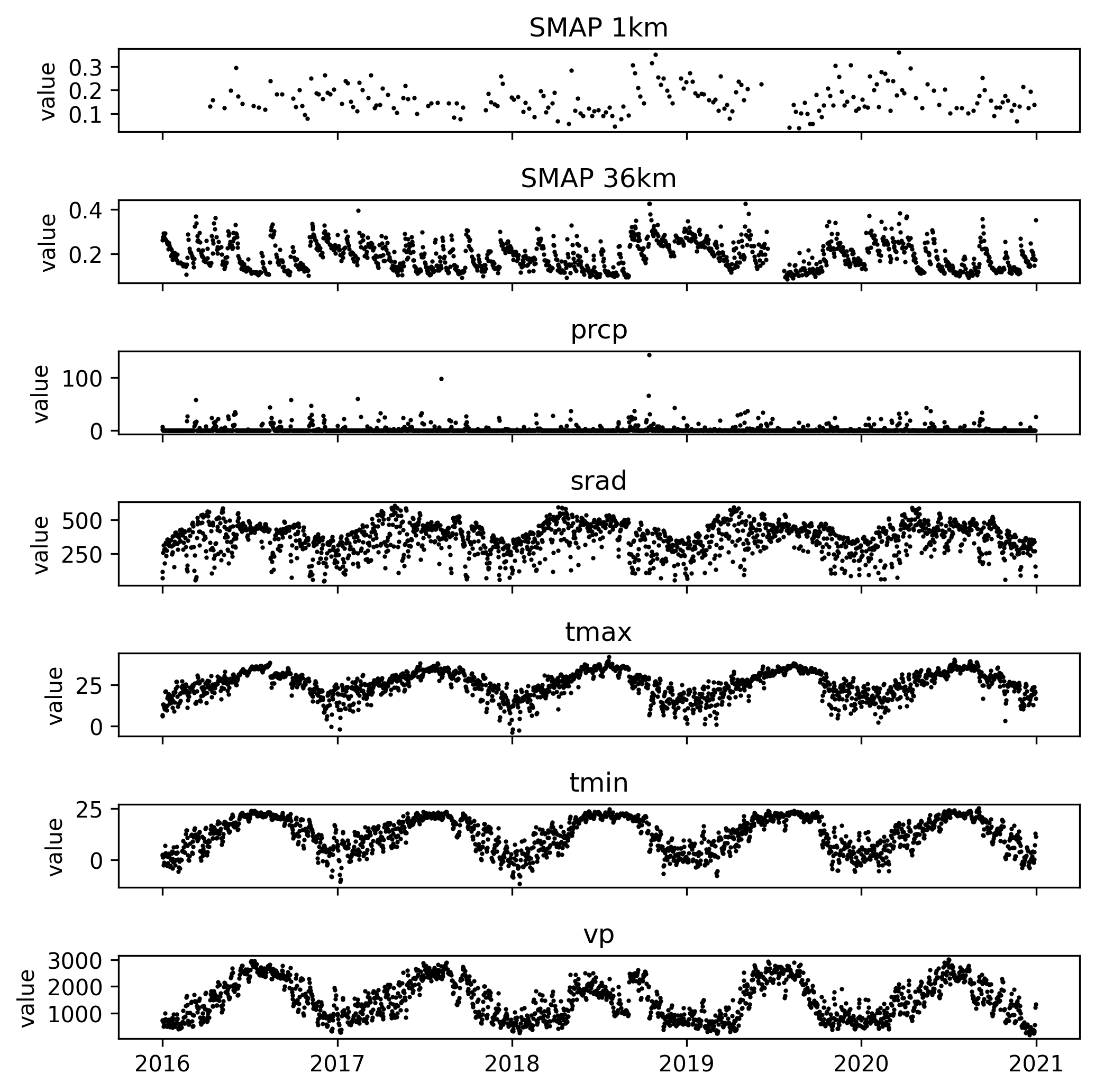}
\caption{SMAP 1km, SMAP 36km, and other daily surface weather data at a 1km × 1km location over the period from January 1, 2016, to December 31, 2020. The weather data include prcp (precipitation), srad (solar radiation), tmax (maximum temperature), tmin (minimum temperature), and vp (vapor pressure).}
\label{fig: eda_time_varying}
\end{figure}

Figure \ref{fig: eda_time_varying} displays the SMAP 1km, SMAP 36km, and daily surface weather data at a 1km $\times$ 1km location over the period from January 1, 2016, to December 31, 2020. 

\subsection{Partitioning Data for Training, Validation, and Testing}
Let $S$ represent our spatial domain, which is a \(36 \times 36\) km grid, and $T$ our temporal domain, spanning from 2016/01/01 to 2022/12/31.
This configuration yields 1296 spatial locations and 2557 time points. We initially divide all the data temporally into two segments. The segment from 2016/01/01 to 2020/12/31 is allocated for training and validation, while the segment from 2021/01/01 to 2022/12/31 is reserved for out-of-sample testing.

\begin{figure}[H]
     \centering
      \begin{subfigure}[b]{0.45\textwidth}
		\centering
		\includegraphics[width=\textwidth]{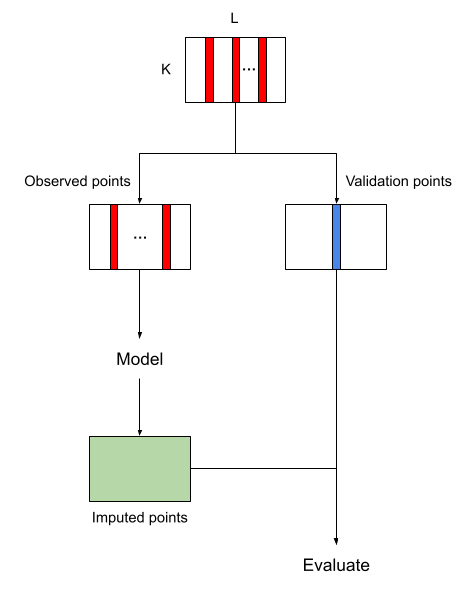}
		\caption{}
		\label{fig: missing at time points}
	 \end{subfigure}
         \hfill
      \begin{subfigure}[b]{0.45\textwidth}
         \includegraphics[width=\textwidth]{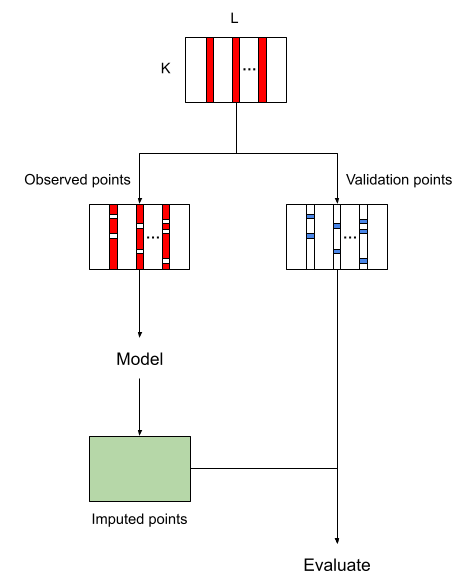}
		 \caption{}
	\label{fig: missing at random}
     \end{subfigure}
    
     \caption{The image on the left describes the ``missing not at random" approach. The image on the right describes the ``missing completely at random" approach. In both figures, white represents missing data, red represents observed data, blue represents validation data, and green represents imputed data.}
     \label{fig: missing pattern}
\end{figure}

In both segments, artificial missingness is injected in SMAP 1km data to facilitate model validation.  We use two distinct approaches for setting missing data: one being ``missing not at random (MNAR)", and the other ``missing completely at random (MCAR)". In the former approach, we randomly select a proportion $p$ of time points, marking the data at all locations for these time points as missing. These points are used for validation. This approach to validation data selection aims to closely mirror the actual missingness patterns observed in real-world data, ensuring a meaningful reflection of model performance on unseen data. On the other hand, the later approach randomly selects a proportion $p$ of points and marks them as missing. This approach seeks to establish more general missing patterns in a spatiotemporal dataset. For both approaches, $p$ is set to 0.2. Figure \ref{fig: missing pattern} describes how these two approaches work.

\subsection{Imputation via Spatiotemporal Transformer Network}\label{sec: Imputation via Spatiotemporal Transformer Network}

\subsubsection*{Initial Setup}\label{sec: Model Training Setup}

Let $\boldsymbol{Y}_{\text{train}}$ denote the training dataset of SMAP 1km, with its corresponding missing mask represented by $\boldsymbol{M}_{\text{train}}$. The predictive variables, including SMAP 36km, Daily Surface Weather Data, Soil, Topography, and Land cover type data, are denoted by $\boldsymbol{X}_{\text{train}}$. In this setup, the total number of spatial locations is 1296, the total number of time points is 1827, and the total number of predictive variables is 15.

It's important to note that $\boldsymbol{X}_{\text{train}}$ may also contain missing values. To tackle this, we create a binary tensor $\boldsymbol{Z}_{\text{train}}$ with the same shape of $\boldsymbol{X}_{\text{train}}$, where $\boldsymbol{Z}_{\text{train}}$ pinpoints the missing positions within $\boldsymbol{X}_{\text{train}}$. Then we concatenate $\boldsymbol{X}_{\text{train}}$ and $\boldsymbol{Z}_{\text{train}}$ along the feature axis, yielding an updated version of $\boldsymbol{X}_{\text{train}}$. As a result, the number of predictive variables becomes 30.

\begin{figure}
\centering
\includegraphics[width=0.8\textwidth]{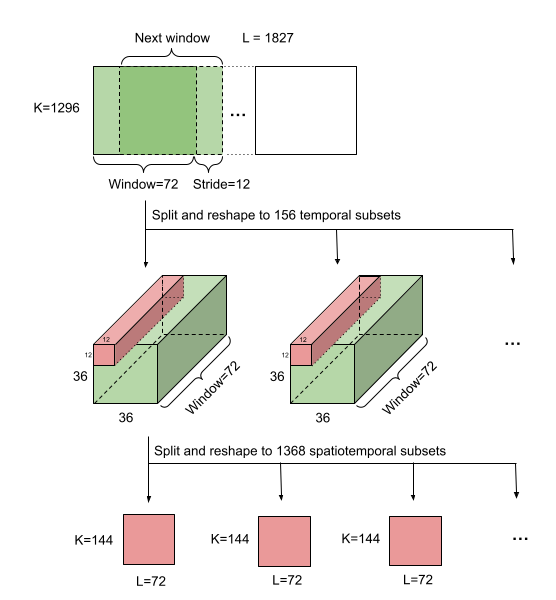}
\caption{ The process of segmenting training data into multiple training samples.}
\label{fig: data_split}
\end{figure}

Next, we divide the training data into multiple training samples. According to \citet{orth2012analysis}, soil moisture memory can last up to 40 days. Therefore, we partition $\boldsymbol{Y}_{\text{train}}$ along the temporal dimension using sliding windows with a fixed length of 72 and a stride of 12.  The chosen window length ensures that the model has access to a sufficient range of temporal data to capture the relevant short-term patterns in soil moisture without unnecessarily extending the memory window to periods where the relevance of data has decayed. This operation results in 152 temporal subsets, as illustrated in Figure \ref{fig: data_split}.

\begin{figure}
\centering
\includegraphics[width=0.5\textwidth]{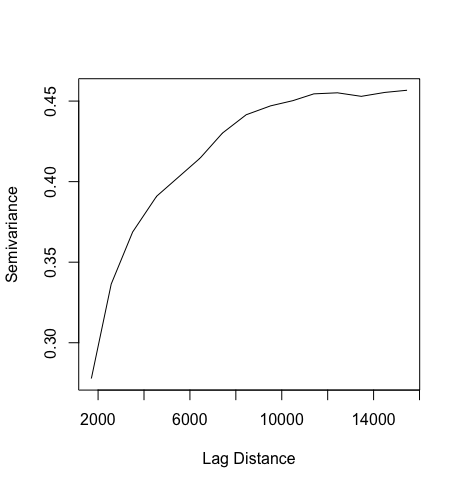}
\caption{The semi-variogram derived from the analysis of spatial residuals. The variance stabilizes at a lag distance of approximately 12 kilometers.}
\label{fig: variogram}
\end{figure}

In determining the appropriate split size for the spatial grid, several factors were taken into consideration. Firstly, a grid size that is too small might result in the loss of crucial spatial correlations between points. Conversely, an overly large grid size, while potentially capturing spatial correlations, tends to include distant points that may not be spatially correlated, and significantly increases the computational cost. Therefore, a balanced grid size is needed. To determine a reasonable grid size, we use a variogram, a tool commonly used to quantify spatial correlations. Here, a linear model was fitted at each location using the training data, from which the residuals were computed. Next, a sample variogram was derived from these residuals to assess the spatial correlations, see Figure \ref{fig: variogram}. By analyzing the variogram, we identified that the variance stabilizes when the distance reaches 12 km, indicating a level where the spatial correlation between the furthest points within a grid becomes minimal. Following this, we further segment each of these temporal subsets along the spatial axis by subdividing the 36x36 km grid into 9 equally sized 12x12 km grids, illustrated in Figure \ref{fig: data_split}. 

This entire process yields a total of 1368 spatiotemporal subsets. The same procedure is replicated for $\boldsymbol{X}_{\text{train}}$ and $\boldsymbol{M}_{\text{train}}$. Consequently, we have 1368 training samples, where each sample consists of $\boldsymbol{Y}_i \in \mathbb{R}^{144\times 72}$, $\boldsymbol{X}_i \in \mathbb{R}^{144 \times 72 \times 30}$, and $\boldsymbol{M}_i \in \mathbb{R}^{144 \times 72}$. The purpose of segmentation is twofold. First, it functions as a data augmentation technique, providing more samples for model training. Second, it enhances the computational efficiency of the transformer model by mitigating the complexity associated with managing long sequences in the attention mechanism.

\subsubsection*{Model Training}
For each training sample, we introduce artificial missing values to $\boldsymbol{Y}_i$ using the same two approaches as used in obtaining validation data. In the case where validation data are ``missing not at random", we randomly select $p_i \in \{0.2, 0.5, 0.8\}$, and mark $p_i$ proportion of time points in $\boldsymbol{Y}_i$ as missing. In the ``missing completely at random" scenario, we randomly mark $p_i$ proportion of entries in $\boldsymbol{Y}_i$ as missing. This procedure results in different positions and proportions of missingness across training samples, enabling the model to learn from a diverse set of scenarios and prevent overfitting.

\subsubsection*{Model Configuration}
In our model, $\text{MLP}$ is a feed-forward network comprising two linear transformations with a ReLU activation function in between, as expressed below:
\begin{equation}
    \text{MLP}(x)= \max (0, xW_1+b_1)W_2+b_2,
\end{equation}

Within the input encoder, both the hidden and output dimensions of the MLP are set to 64, alongside a spatial and temporal encoding dimension of 64. 

For the spatiotemporal transformer encoder, we use four spatiotemporal attention layers. Within each layer, the temporal attention layer uses a single-layer MSA with one attention head, while the spatial layer is explored in two variants: one with a single-layer MSA, and the other with two layers of SW-MSA. In the latter, the first layer uses a window size of $4\times 4$ with no shift (shift-size of 0), and the second layer uses a window size of $4 \times 4$ with a shift-size of 2. The hidden dimension of the MLP within the encoder is set to 64. 

In the output layer, we set the hidden dimension to 64 and the output dimension to 1 for the MLP. 

As for hyperparameters, we set the batch size as 16 and, the number of training epochs to 200. For optimization, We use Adam with a cosine scheduled learning rate, tapering from 0.001 down to 0.0001 over the training period \citep{kingma2014adam}.

\subsubsection*{Model Evaluation}
After model training, the model's performance is assessed through the imputation of the validation points. During evaluation, the training dataset is again divided into smaller subsets, akin to the training phase. However, unlike the training phase, no additional masks are applied to $\boldsymbol{Y}_i$ during evaluation. After obtaining imputations for all training samples, we reverse the splitting process to reconstruct $\boldsymbol{Y}_{\text{train}}$. In cases where multiple imputations are made at a specific point in the original data, we take the average. Finally, we evaluate the imputation performance using the mean absolute error (MAE) and mean relative error (MRE). Suppose $y_i$ is the ground truth of $i$th validation point, $\hat{y}_i$ is the corresponding imputed value, and $N$ is the number of validation points in total. Then MAE and MRE are defined as:

\begin{align}
		\text{MAE}=\sum_{i=1}^N |\hat{y}_i - y_i|/N\\
		\text{MRE}=\sum_{i=1}^N|\hat{y}_i-y_i|/{\sum_{i=1}^N|y_i|}
\end{align}

Next, we apply the same procedure to assess the model's performance on the test data. Since the test data is untouched during training, it serves as a reliable measure of the model's generalization ability.

\subsection{Comparison with Baseline Methods}
Our model is benchmarked against various simple and advanced imputation methods. These methods include:

\begin{description}
	\item \textbf{Mean}: For each location, missing values are replaced with the monthly mean.
	\item \textbf{Linear Interpolation}: Missing values are estimated by drawing a straight line between two known values and finding the value at the missing point along this line.
	\item \textbf{Matrix Factorization}: Matrix Factorization is used to impute missing values by decomposing the original data matrix into two lower-dimensional matrices, capturing underlying patterns and structures. The missing values are then estimated by reconstructing the data matrix from these factorized matrices.
	\item \textbf{Mice}: Multiple imputation by chained equations (MICE) works by performing multiple imputations through a series of regression models, each time estimating missing values, thus creating multiple imputed datasets \citep{white2011multiple}. The final imputation could be an average of these datasets.
	\item \textbf{ImputeTS}:  ImputeTS is an R package for missing value imputation, which utilizes the state space model and Kalman smoothing \citep{moritz2017imputets}.
	\item \textbf{Random Forest}: Random Forest utilizes an ensemble of decision trees to predict missing values, considering each missing entry as a target variable while treating external covariates as features \citep{breiman2001random}.
	\item \textbf{GRIN}: Graph Recurrent Imputation Network (GRIN) uses a Graph-based Bidirectional Recurrent Neural Network to capture spatial and temporal dependencies in data, leveraging these relationships to impute missing values \citep{cini2021filling}.
	\item \textbf{CSDI}: Conditional Score-based Diffusion models for Imputation (CSDI) introduce a conditional score-based diffusion model for probabilistic imputation \citep{tashiro2021csdi}.
\end{description}

Among the discussed methods, Mean, Linear Interpolation, Matrix Factorization, and ImputeTS do not utilize any predictive variables. In contrast, with minor modifications, the GRIN and CSDI models can incorporate covariates to assist in imputation. It's also worth noting that methods such as Mean, Matrix Factorization, MICE, and Random Forest solely rely on predictive variables to impute missing values, disregarding the inherent time series structural information. However, other methods do consider the time series information in the imputation process.

Additionally, methods such as MF and MICE do not offer a general model form for imputation, which requires retraining when applied to unseen datasets, thereby limiting their generalizability.

\subsection{Results}
\subsubsection*{Missing Not at Random}
Initially, we focus on data imputation where validation points are ``missing not at random". For robustness, each imputation algorithm is independently executed five times. During each run, the selection of validation points is randomized, which may lead to variability in chosen points across runs.

The imputation results are presented in Table \ref{tab: missing at time points}. Our ST-Transformer models outperform the baseline models in both validation and testing datasets. Specifically, the model using SW-MSA for spatial modeling is slightly better than the one using global attention. Furthermore, models that incorporate covariate information tend to perform better than those that do not, indicating the informative nature of external covariates in imputing SMAP 1km data. Additionally, the deep learning-based methods outperform traditional methods and show greater robustness (as evidenced by smaller standard errors in five replicates). This enhanced performance is likely due to their ability to model complex, non-linear relationships in the data, which traditional imputation methods might fail to capture. 

In terms of training efficiency, the ST-Transformer model demonstrates a comparative advantage over certain other deep learning methods. For instance, when trained on an NVIDIA T4 GPU, the GRIN model requires approximately 8 hours to complete its training process. In contrast, the ST-Transformer model completes its training in about 2 hours. Additionally, while the CSDI model matches the ST-Transformer in training duration, it necessitates an extra 4 hours for generating imputation results. 

\begin{figure}
\centering
\includegraphics[width=\textwidth]{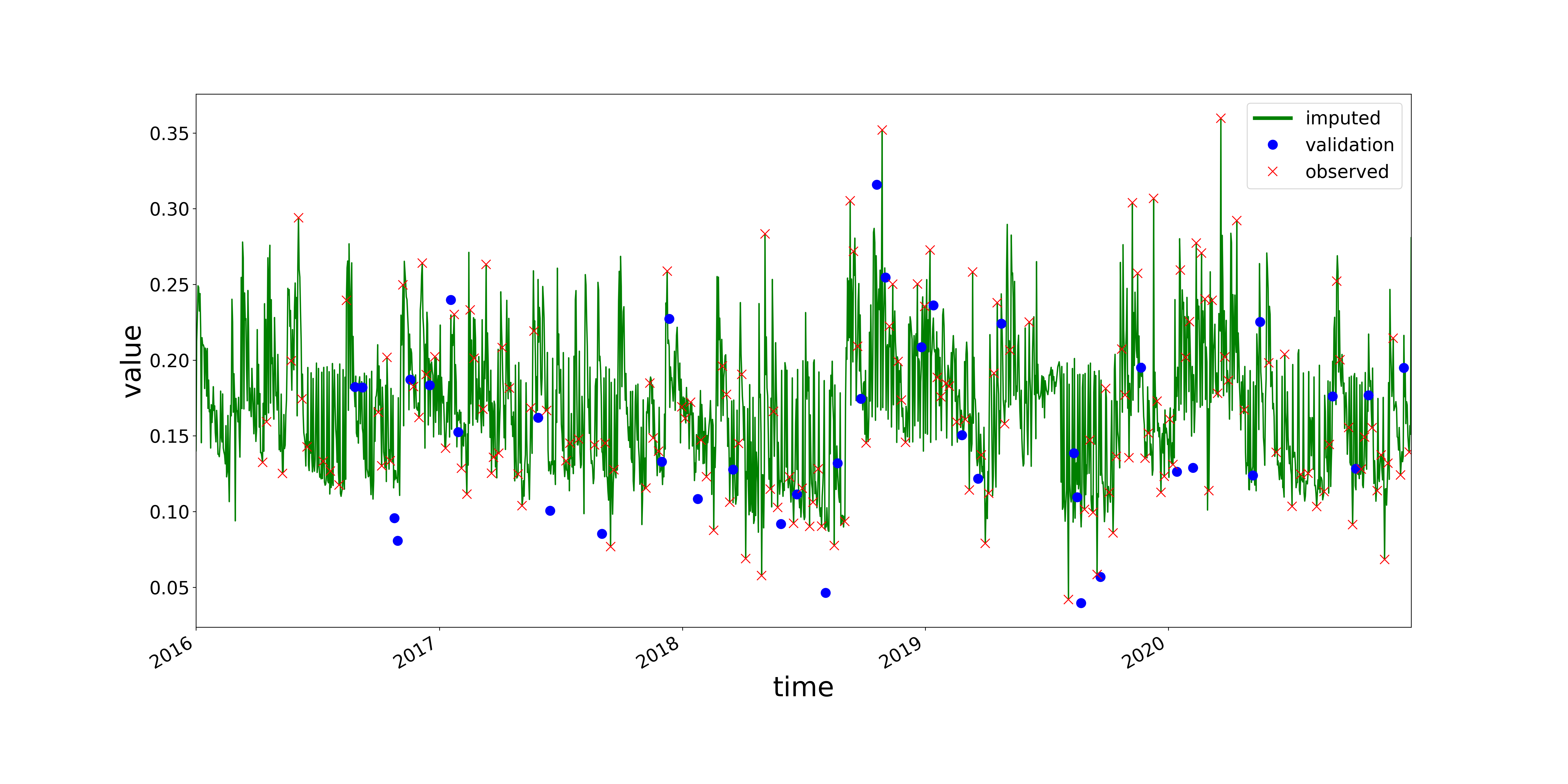}
\caption{Imputation of SMAP 1km from 2021-01-01 to 2022-12-31 at a 1km $\times$ 1km location. The land cover is grassland. Here, the green line is the imputed soil moisture data, red $\times$ represents the observed data, and blue $\circ$ represents the validation data.}
\label{fig: imputation_result}
\end{figure}

\begin{figure}[H]
\centering
\includegraphics[width=\textwidth]{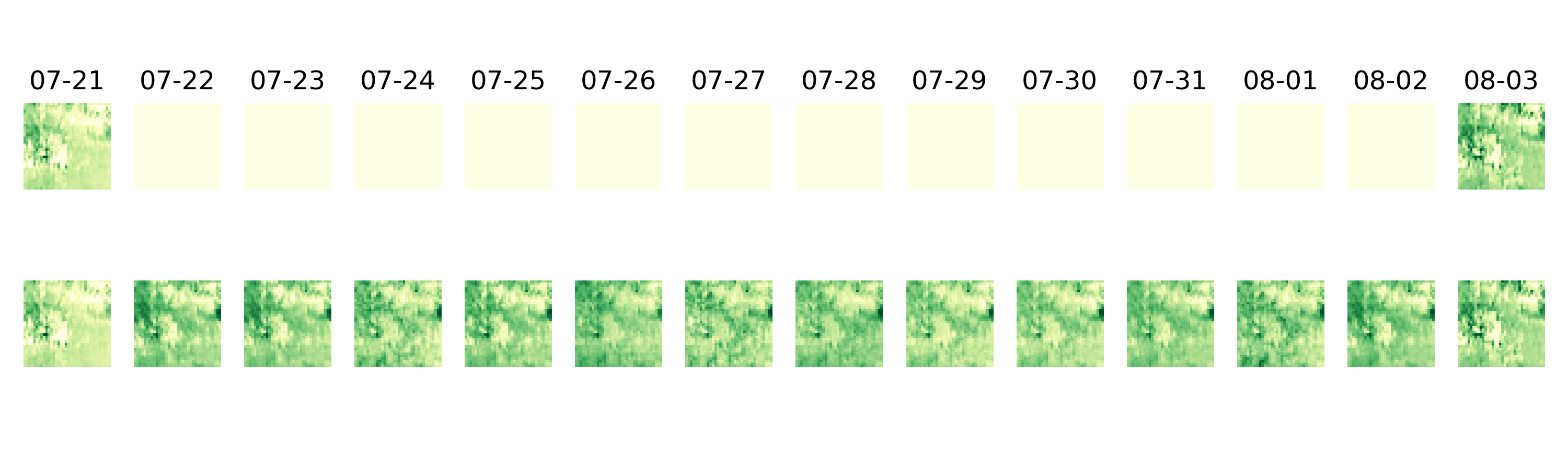}
\caption{Imputation of SMAP 1km data spanning from July 21, 2021, to August 3, 2021. The top row displays the observed SMAP 1km data, where white spaces represent missing data. The bottom row illustrates the imputed soil moisture data.}
\label{fig: missing_at_time_points_smap} 
\end{figure}

Figure \ref{fig: imputation_result} displays the imputation performance of the ST-Transformer (SW-MSA) at a 1km $\times$ 1km location in the test dataset.  Figure \ref{fig: missing_at_time_points_smap} presents the imputation results across all locations within the study region over a span of 14 days from July 21, 2021, to August 3, 2021.

\begin{table}[h!]
    \centering
    \begin{tabularx}{\textwidth}{llXX}
        \toprule
        Dataset & Method & MAE & MRE \\
        \midrule
        \multirow{10}{*}{Validation}&
        Mean & 0.0473 +/- 0.001& 31.12\% +/- 2.10\%  \\
        &Linear Interpolation & 0.0484 +/- 0.000 & 32.55\% +/- 0.32\%\\
        &Matrix Factorization & 0.0491+/- 0.003 & 43.25\% +/- 6.54\% \\
        &ImputeTS & 0.0413 +/- 0.002 & 27.63\% +/- 4.83\% \\
        &Mice & 0.0312 +/- 0.001& 27.32\% +/- 2.36\% \\
        &Random Forest &  0.0316 +/- 0.001&  27.25\% +/- 2.68\%\\
        &GRIN & 0.0287 +/- 0.000 & 16.76\% +/- 0.42\%\\
        &CSDI & 0.0283 +/- 0.000 &16.85\% +/- 0.50\%\\
        &ST-Transformer (MSA) & 0.0240 +/- 0.000 & 15.14\% +/- 0.59\%\\
        &\textbf{ST-Transformer (SW-MSA)} & 0.0223 +/- 0.000 & 14.66\% +/- 0.63\%\\
        
        \midrule
        \multirow{10}{*}{Testing}&Mean & 0.0482 +/- 0.000 & 32.24\% +/- 0.91\%  \\
        &Linear Interpolation & 0.0485 +/- 0.000& 32.59\% +/- 0.30\%\\
        &Matrix Factorization & 0.0408 +/- 0.005& 34.22\% +/- 4.64\% \\
        &ImputeTS & 0.0415 +/- 0.003 & 25.11\% +/- 3.30\% \\
        &Mice & 0.0325 +/- 0.001& 28.10\% +/- 2.27\% \\
        &Random Forest &  0.0344 +/- 0.001&  30.81\% +/- 1.86\%\\
        &GRIN & 0.0289 +/- 0.000 & 17.10\% +/- 0.49\%\\
        &CSDI & 0.0284 +/- 0.000 &17.51\% +/- 0.65\%\\
        &ST-Transformer (MSA) & 0.0247 +/- 0.001 & 16.62\% +/- 0.87\%\\

        &\textbf{ST-Transformer (SW-MSA)} & 0.0231 +/- 0.001 & 15.51\% +/- 0.91\%\\
        \bottomrule

    \end{tabularx}
    \caption{Average performance of imputation methods over five independent runs. The data follows a ``missing not at random" pattern. The table compares various imputation algorithms.}
    \label{tab: missing at time points}
\end{table}

\subsubsection*{Missing Completely at Random}
The ``missing not at random" scenario is less common in spatiotemporal data. Typically, missing values in spatiotemporal data are distributed randomly across space and time. To ensure the generalizability of our method to these situations, we also evaluate all imputation methods on the ``missing completely at random" scenario. Similar to the earlier scenario, every imputation algorithm is run five times independently. The outcomes are presented in Table \ref{tab: missing at random}. Our ST-Transformer models consistently outperform all baseline models. Methods like Mean, Linear interpolation, ImputeTS, Mice, and Random Forest, which do not model spatial correlation, exhibit relatively stable performance. Interestingly, Matrix Factorization outperforms the covariate-based methods and even matches the performance of some deep learning methods. This suggests the presence of non-missing data at certain spatial locations significantly aids the imputation process. Additionally, our model with the SW-MSA layer significantly outperforms the model using the standard MSA layer. This indicates that the SW-MSA layer is more effective in capturing spatial correlations compared to the global attention approach.

\begin{table}[h!]
    \centering
    \begin{tabularx}{\textwidth}{llXX}
        \toprule
        Dataset & Method & MAE & MRE \\
        \midrule
         \multirow{10}{*}{Validation}&Mean & 0.0483 +/- 0.000& 32.13\% +/- 0.85\%  \\
        &Linear Interpolation & 0.0466 +/- 0.000 & 30.98\% +/- 0.094\%\\
        &Matrix Factorization & 0.0240+/- 0.000 & 19.33\% +/- 0.12\% \\
        &ImputeTS & 0.0410 +/- 0.002 & 27.12\% +/- 0.92\% \\
        &Mice & 0.0226 +/- 0.001& 18.61\% +/- 0.17\% \\
        &Random Forest &  0.0319 +/- 0.001&  26.55\% +/- 0.21\%\\
   		&GRIN &0.215 +/- 0.000 & 14.78\% +/- 0.43\%\\
        &CSDI & 0.211 +/- 0.000 & 14.01\% +/- 0.48\%\\
        &ST-Transformer (MSA) & 0.0192 +/- 0.000 & 12.76\% +/- 0.74\%\\
        &\textbf{ST-Transformer (SW-MSA)} & 0.0144 +/- 0.000 & 9.58\% +/- 0.34\%\\
        
        \midrule
        \multirow{10}{*}{Testing}&Mean & 0.0484 +/- 0.000 & 32.55\% +/- 0.00\%  \\
        &Linear Interpolation & 0.0467 +/- 0.000& 31.45\% +/- 0.00\%\\
        &Matrix Factorization & 0.0240 +/- 0.000& 19.27\% +/- 0.13\% \\
        &ImputeTS & 0.0415 +/- 0.003 & 25.11\% +/- 3.25\% \\
        &Mice & 0.0230 +/- 0.000& 19.07\% +/- 0.17\% \\
        &Random Forest &  0.0364 +/- 0.000&  32.91\% +/- 0.30\%\\
        &GRIN &0.0220 +/- 0.000 & 15.31\% +/- 0.22\%\\
        &CSDI & 0.0214 +/- 0.000 & 14.96\% +/- 0.51\%\\
        &ST-Transformer (MSA) & 0.0195 +/- 0.001 & 13.16\% +/- 0.40\%\\

        &\textbf{ST-Transformer (SW-MSA)} & 0.0146 +/- 0.001 & 9.80\% +/- 0.18\%\\
        
        \bottomrule

    \end{tabularx}
    \caption{Average performance of imputation methods over five independent runs. The data follows a ``missing completely at random" pattern. The table compares various imputation algorithms.}
    \label{tab: missing at random}
\end{table}

\subsection{Ablation Studies}

To highlight the effectiveness of spatiotemporal attention, we conduct an ablation study comparing four model variations: the complete ST-Transformer, a variant without spatial or temporal attention, one with only temporal attention, and another with only spatial attention. We evaluate these models on the validation data under both ``missing not at random" and ``missing completely at random" scenarios, with the results presented in Table \ref{tab: ablation_study_1}. In both scenarios, incorporating temporal and spatial attention significantly enhances imputation accuracy. Interestingly, while spatial attention doesn't significantly improve results in the ``missing not at random" scenario, it brings substantial improvement in the ``missing at random" scenario. One reason is that in the former scenario, observations are missing for all locations, leaving spatial information unexploited. Conversely, in the latter scenario, the model can directly leverage observed data from nearby locations for imputation.

\begin{table}[h!]
    \centering
    \begin{tabularx}{\textwidth}{XXXX}
        \toprule
         Scenario&Method & MAE & MRE \\
        \midrule
        \multirow{4}{*}{MNAR}&\textbf{ST-Transformer} & 0.0223 +/- 0.000 & 14.66\% +/- 0.63\%  \\
        &ST-Transformer (no space, no time) & 0.0301 +/- 0.000& 19.89\% +/- 0.73\%\\
        &ST-Transformer (no space) & 0.0257 +/- 0.000& 17.17\% +/- 0.62\% \\
        &ST-Transformer (no time) & 0.0262 +/- 0.000 & 17.89\% +/- 0.15\% \\
        \midrule
        \multirow{4}{*}{MCAR}&\textbf{ST-Transformer} & 0.0144 +/- 0.000 & 9.58\% +/- 0.34\%  \\
        &ST-Transformer (no space, no time) & 0.0300 +/- 0.000& 19.80\% +/- 0.11\%\\
        &ST-Transformer (no space) & 0.0213 +/- 0.000& 14.17\% +/- 0.42\% \\
        &ST-Transformer (no time) & 0.0171 +/- 0.000 & 11.50\% +/- 0.11\% \\
        \bottomrule

    \end{tabularx}
    \caption{Average imputation performance over five independent runs at two missing scenarios. The models compared are ST-Transformers with different structures.}
    \label{tab: ablation_study_1}
\end{table}

Additionally, we conducted another ablation study to show the impact of covariates on imputation. We categorize the covariates into two groups: time-varying features, which include SMAP 36km and daily surface weather data, and static features like soil characteristics, topography, and land cover type data.

We evaluate four model variations. The first model incorporates both time-varying and static covariates, the second model includes no covariates, the third model only uses time-varying features, and the fourth model solely uses static features. We assess these model variations under both missing data scenarios.

Table \ref{tab: ablation_study_2} presents the imputation performance. In the ``missing not at random" scenario, we observe that including time-varying covariates significantly enhances the model performance, while incorporating static covariates also provides a slight improvement. Conversely, in the ``missing completely at random" scenario, the inclusion of covariates doesn't contribute much to the performance. This is probably because the model imputes missing values by exploiting spatial relations, making the additional covariate information less influential.

\begin{table}[h!]
    \centering
    \begin{tabularx}{\textwidth}{XXXX}
        \toprule
        Scenario&Method & MAE & MRE \\
        \midrule
        \multirow{4}{*}{MNAR}&\textbf{ST-Transformer} &  0.0223 +/- 0.000 & 14.66\% +/- 0.63\%  \\

        &ST-Transformer (no covariates) & 0.0404 +/- 0.001& 24.75\% +/- 1.51\%\\
        &ST-Transformer (time-varying) & 0.0254	 +/- 0.001& 16.17\% +/- 0.66\% \\
        &ST-Transformer (static) & 0.0385 +/- 0.001 & 22.39\% +/- 0.50\% \\
        \midrule
          \multirow{4}{*}{MCAR}&\textbf{ST-Transformer} & 0.0144 +/- 0.000 & 9.58\% +/- 0.34\%  \\
        &ST-Transformer (no covariates) & 0.0155 +/- 0.001& 10.31\% +/- 0.38\%\\
        &ST-Transformer (time-varying) & 0.0147 +/- 0.001& 9.79\% +/- 0.11\% \\
        &ST-Transformer (static) & 0.0150 +/- 0.001 & 9.93\% +/- 0.15\% \\
        \bottomrule

    \end{tabularx}
    \caption{Average imputation performance over five independent runs at two missing scenarios. The models compared are ST-Transformers with different covariates.}
    \label{tab: ablation_study_2}
\end{table}

\subsection{Spatial variability of the Imputation}
\begin{figure}
     \centering
      \begin{subfigure}[b]{0.45\textwidth}
		\centering
		\includegraphics[width=\textwidth]{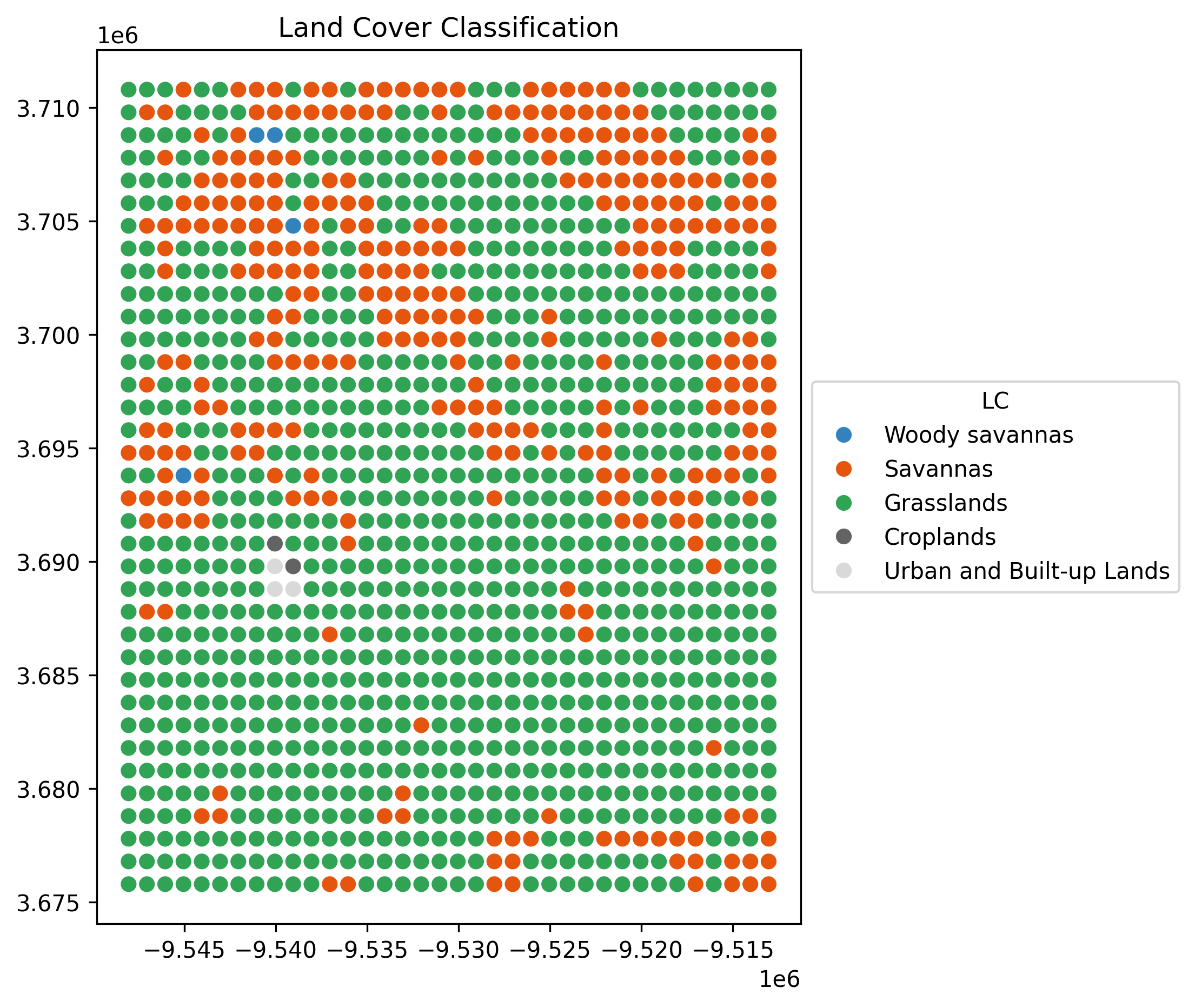}
		\caption{}
		\label{fig: missing at time points}
	 \end{subfigure}
         \hfill
      \begin{subfigure}[b]{0.45\textwidth}
         \includegraphics[width=\textwidth]{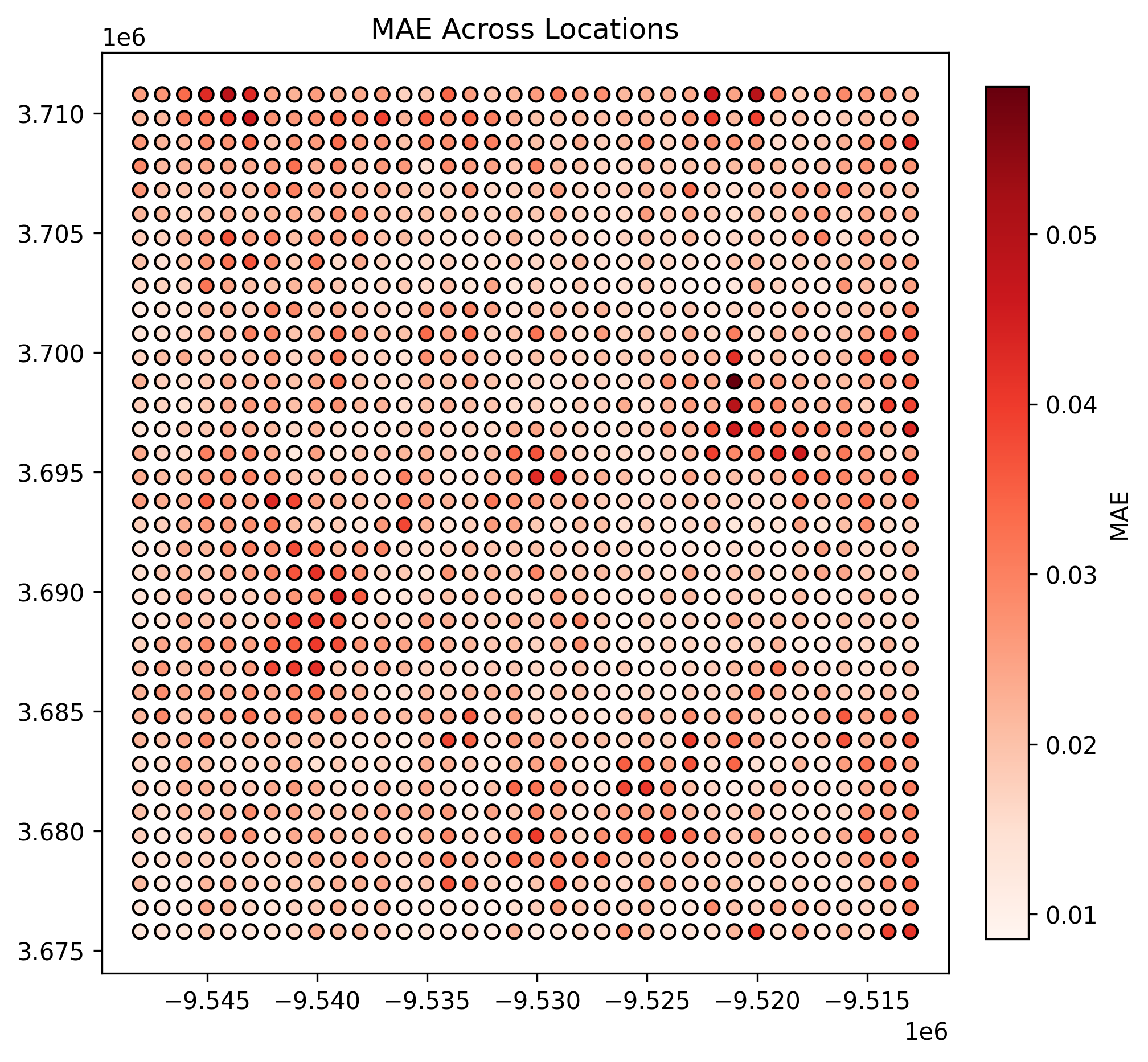}
		 \caption{}
	\label{fig: missing at random}
     \end{subfigure}
     
    \begin{subfigure}[b]{\textwidth}
	\includegraphics[width=\textwidth]{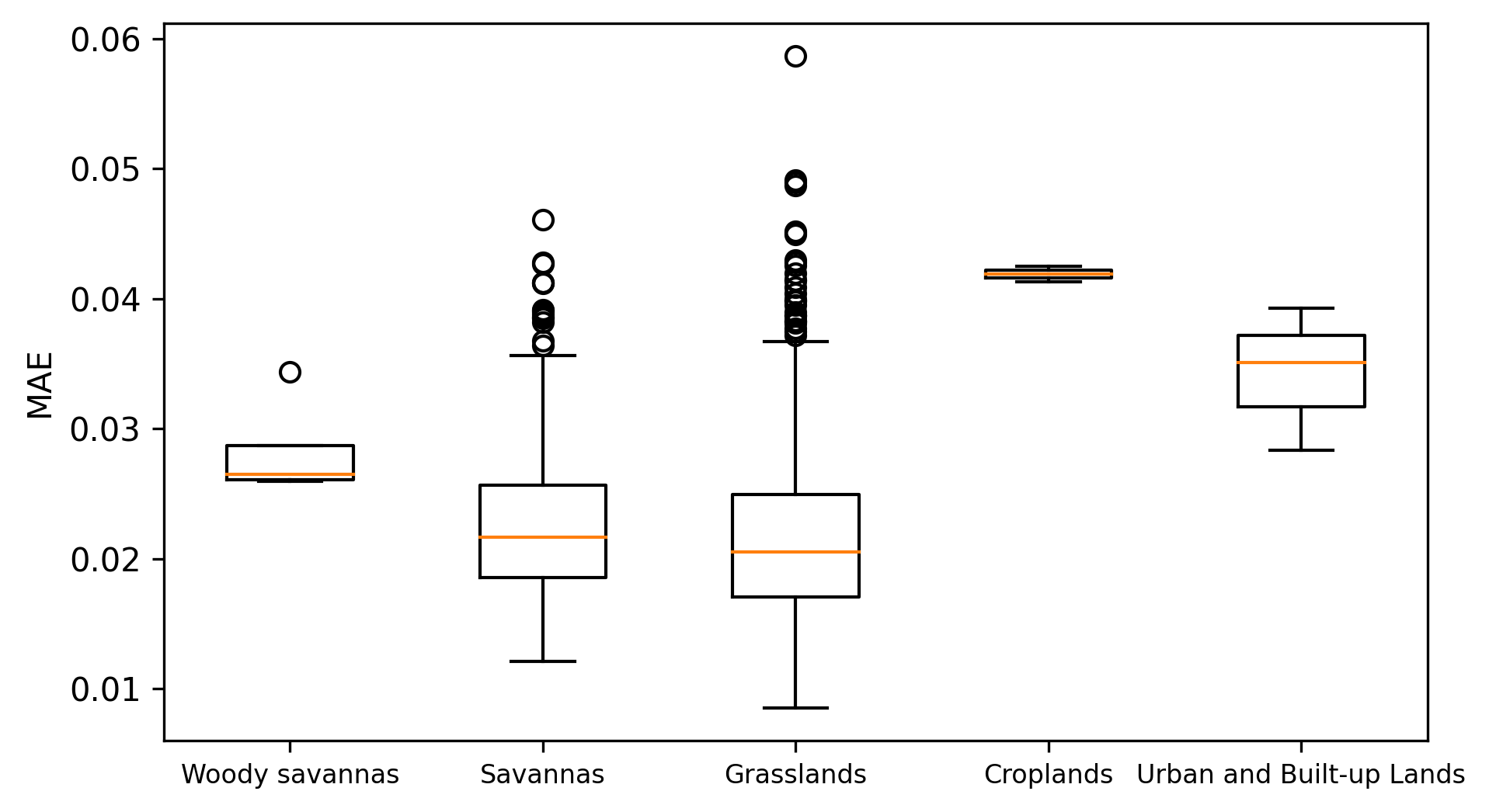}
	\end{subfigure}
    
     \caption{Figure (a) illustrates the spatial distribution of the land cover types. Figure (b) shows the spatial variation of the MAE. Figure (c) provides box plots of the MAE across different land cover types.}
     \label{fig: mae_across_space}
\end{figure}

In our study, we evaluated the imputation accuracy of different land covers by measuring the Mean Absolute Error (MAE) for various locations. Under the ``missing not at random" scenario, we plot the MAE across space in Figure \ref{fig: mae_across_space}. We observed distinct variations in imputation accuracy among different land covers. Specifically, grasslands showed the best performance, followed by savannas, woody savannas, urban areas, and croplands. The poorer imputation in cropland and urban areas could be attributed to a scarcity of training data.

\section{Conclusion}
In this paper, we introduced the Spatiotemporal Transformer (ST-Transformer) model, designed to address the challenge of imputing missing values in sparse soil moisture datasets. Compared with traditional soil moisture methods, our model has two key advantages. Firstly, it incorporates both time-varying and static covariates to guide imputation in cases of extreme data sparsity, offering a more robust approach to addressing missing soil moisture data. Secondly, it employs a spatiotemporal transformer to effectively model spatial and temporal correlations from the available data. The utilization of SW-MSA layers allows the model to efficiently handle large spatial fields. 

Furthermore, we established a self-supervised training framework and demonstrated its efficacy in imputing SMAP 1km data. The model's ability to generalize to other spatiotemporal imputation tasks was also investigated. In A's simulation study (shown in the Appendix), the model was applied to the Healing Mnist dataset, where it surpassed the performance of the current leading benchmark. In both simulation studies and real data analysis, our ST-Transformer model showed enhanced accuracy and computational efficiency compared to existing imputation methods.

\bibliographystyle{apalike}
\bibliography{main}

\appendix

\section{Simulation} 
To assess the effectiveness of our ST-Transformer model, we also conducted two distinct simulation studies. The first involved imputation tasks on the Healing MNIST dataset, and the second focused on imputation using the Soil Moisture Hydroblocks dataset.

\subsection{Healing MNIST}

The Healing MNIST dataset \citep{krishnan2015deep} is a variant of the popular MNIST dataset \citep{lecun1998gradient}. The original MNIST dataset comprises a large collection of handwritten digits (0 through 9), each represented as a 28x28 pixel grayscale image. The Healing MNIST dataset adds a dynamic aspect to these images, transforming them into time series or sequential data. This transformation is typically achieved by applying various changes to the digits like rotation, scaling, or morphing—across a series of frames, effectively creating a ``video" of each digit undergoing transitions. For testing our imputation model, we intentionally introduced missing values into each frame and then utilized the remaining observed data to predict these missing elements. Following the approach of \citet{fortuin2020gp}, we generated missing values in two ways: missing completely at random (MCAR), and missing not at random (MNAR). In the latter scenario, the likelihood of a white pixel becoming missing was 2 times higher than that of a black pixel. In both scenarios, approximately 50\% of each frame's pixels were masked as missing. The dataset comprised 50,000 time series of digits for training purposes and 10,000 for testing. 

\begin{figure}[H]
\centering
\includegraphics[width=\textwidth]{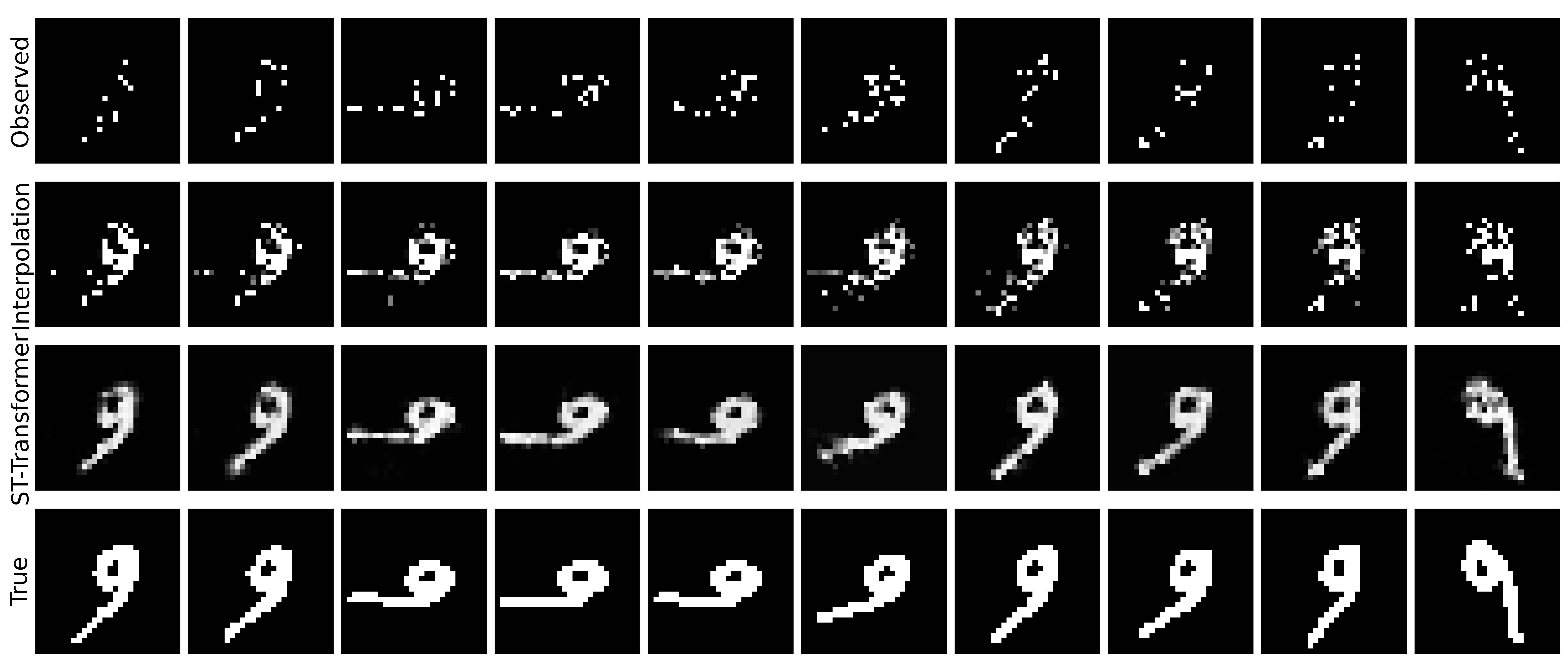}
\caption{Reconstructions from Healing MNIST in the missing not at random scenario. The first row displays a time series of digits with missing values. The second row shows the imputation result using linear interpolation. The third row shows the imputation result using ST-Transformer. The fourth row shows the true sequences of the digits.}
\label{fig: healing_mnist}
\end{figure}

We evaluated our model’s performance against the GP-VAE model \citep{fortuin2020gp} model, which is the state-of-the-art for this type of task, along with basic methods like mean and interpolation. Attempts were made to implement other deep learning methods such as GRIN and CSDI. However, these were found to be excessively demanding in terms of computational resources, even during initial test runs. Consequently, we decided not to include them in our analysis. Our results, detailed in Table \ref{tab: healing_mnist}, show that the ST-Transformer model significantly outperformed the GP-VAE across both missing data scenarios, particularly in the MNAR scenario where the improvement was substantial. Figure \ref{fig: healing_mnist} displays the imputation results in this scenario, illustrating the model's effectiveness in accurately imputing the missing digits.

\begin{table}[h!]
    \centering
    \begin{tabularx}{\textwidth}{XXX}
        \toprule
        Scenario&Method & MSE \\
        \midrule
        \multirow{2}{*}{MCAR}&\textbf{ST-Transformer} & 0.0250+/-0,000   \\

        &GP-VAE& 0.036+/-0.000\\
        &Mean&0.113 +/- 0.001\\
        &Interpolation&0.074 +/- 0.001\\
        
        \midrule
          \multirow{2}{*}{MNAR}&\textbf{ST-Transformer} & 0.054+/-0.000  \\
        &GP-VAE&0.114+/-0.002\\
        &Mean&0.210+/-0.004\\
        &Interpolation& 0.135+/-0.002\\
        \bottomrule

    \end{tabularx}
    \caption{Average imputation performance on the Healing MNIST dataset over 5 independent runs.}
    \label{tab: healing_mnist}
\end{table}

\subsection{SMAP-HydroBlocks}
SMAP-HydroBlocks (SMAP-HB), introduced in \citet{vergopolan2021smap}, is a high-resolution, satellite-based surface soil moisture dataset covering the conterminous United States from 2015 to 2019 at a 30-m resolution. The dataset includes a post-processed, aggregated version at 1 km and 6-hour resolution, available at \href{https://zenodo.org/records/5206725}{https://zenodo.org/records/5206725}. To ensure consistency with our real data analysis, we extract data from a 36 $\times$ 36 km grid within the study region and incorporate covariates to help in the imputation process. We replicate the missing patterns found in the original SMAP 1km dataset and use only the observed data for imputation. 

\begin{figure}
\centering
\includegraphics[width=\textwidth]{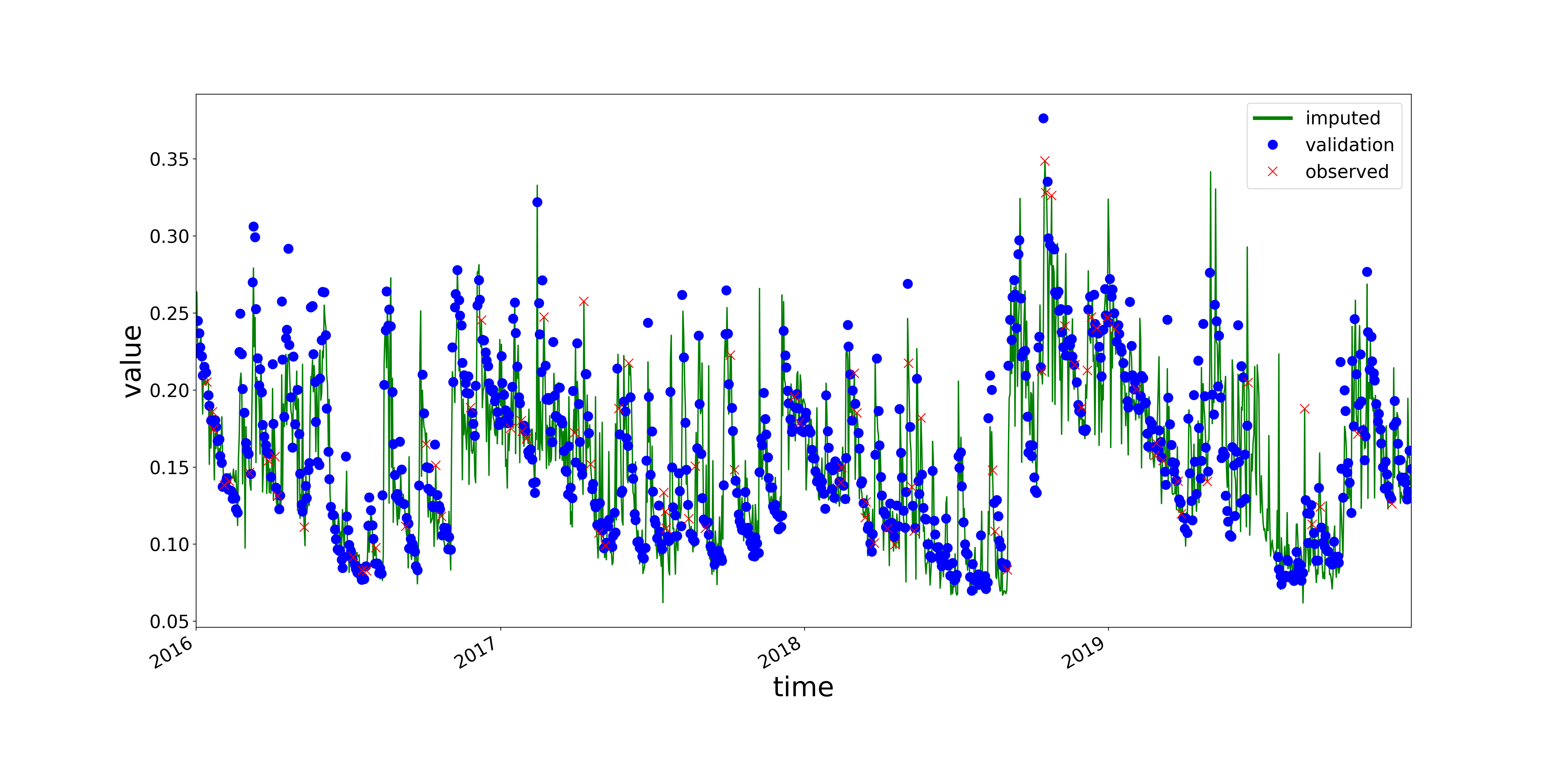}
\caption{Imputation of SMAP HydroBlocks 1km from 2016-01-01 to 2019-12-31 at a 1km $\times$ 1km location using ST-Transformer. The land cover is grassland. Here, the green line is the imputed soil moisture data, red $\times$ represents the observed data, and blue $\circ$ represents the validation data.}
\label{fig: smap_hydroblock}
\end{figure}

The imputation result, presented in Table \ref{tab: smap_hydroblock}, shows that our ST-Transformer model outperforms other methods. Additionally, Figure \ref{fig: smap_hydroblock} provides a visual representation of the imputation results, showing that the data imputed by our model aligns closely with the actual data. 
\begin{table}
    \centering
    \begin{tabularx}{\textwidth}{lXX}
        \toprule
       Method & MAE & MRE \\
        \midrule
        Mean & 0.0456 +/- 0.002& 28.34\% +/- 0.41\%  \\
        Linear Interpolation & 0.0415 +/- 0.001 & 25.16\% +/- 0.080\%\\
        Matrix Factorization & 0.0483+/- 0.000 & 35.29\% +/- 1.36\% \\
        ImputeTS & 0.0410 +/- 0.002 & 27.12\% +/- 0.92\% \\
        Mice & 0.0286 +/- 0.001& 20.09\% +/- 0.60\% \\
        Random Forest &  0.0294 +/- 0.001&  20.35\% +/- 0.397\%\\
   		GRIN &0.259 +/- 0.000 & 16.11\% +/- 0.43\%\\
        CSDI & 0.264 +/- 0.000 & 16.50\% +/- 0.48\%\\
        ST-Transformer (MSA) & 0.024 +/- 0.000 & 15.29\% +/- 0.43\%\\
        \textbf{ST-Transformer (SW-MSA)} & 0.021 +/- 0.000 & 13.17\% +/- 0.24\%\\
        
        \bottomrule

    \end{tabularx}
    \caption{Average imputation performance over 5 independent runs. The data used is from the SMAP HydroBlocks dataset. The table compares various imputation algorithms.}
    \label{tab: smap_hydroblock}
\end{table}

\end{document}